\documentclass[10pt]{article}

\usepackage{acl2002}
\usepackage{amssymb}
\usepackage{graphicx}
\usepackage{alltt}
\usepackage{ifthen}
\usepackage{pstricks}
\usepackage{pst-node}
\usepackage{xspace}

\makeatletter

\title{\vspace{-75pt}
{\normalsize {\it \hfill Appears in  Proc. 2002 Conf. on Empirical Methods in
Natural Language Processing (EMNLP)}} \\ \mbox{}\\Bootstrapping Lexical Choice via Multiple-Sequence Alignment}

\author{
Regina Barzilay \\
Department of Computer Science \\
Columbia University \\
New York, NY  10027 USA\\
\texttt{regina@cs.columbia.edu}
\And
Lillian Lee \\
Department of Computer Science \\
Cornell University \\
Ithaca, NY 14853 USA \\
\texttt{llee@cs.cornell.edu}
}

\newenvironment{frameit}[1]
  {\begin{tabular}{|p{#1}|}\hline}{\\\hline\end{tabular}}

\newcommand{\omt}[1]{}
\newcommand{\set}[1]{\{#1\}}
\newcommand{\semantics}[1]{\texttt{#1}}
\newcommand{\realize}[1]{\textsc{#1}}
\newcommand{\entry}[2]{\semantics{#1} $\rightarrow$ \realize{#2}}
\newcommand{\entrymultiverb}[2]{\semantics{#1}$\rightarrow$ \set{\realize{#2}}}
\newcommand{\twop}{one-parallel\xspace}
\newcommand{\multip}{multi-parallel\xspace}
\newcommand{\msa}{MSA\xspace}
\newcommand{\msas}{{\msa}s\xspace}
\newcommand{\allat}{lattice\xspace}

\newcommand{\allats}{{\allat}s\xspace}

\newcommand{\simfn}{\textrm{sim}}
\newcommand{\alphabet}{\Sigma}

\newcommand{\underscore}{\underline{~}}
\newcommand{\firstphase}{per-instance alignment\xspace}
\newcommand{\Firstphase}{Per-instance alignment\xspace}
\newcommand{\secondphase}{cross-instance alignment\xspace}
\newcommand{\Secondphase}{Cross-instance alignment\xspace}

\newcommand{\slotlat}{slotted lattice\xspace}
\newcommand{\Slotlat}{Slotted lattice\xspace}
\newcommand{\Slotlats}{Slotted lattices\xspace}
\newcommand{\slotlats}{slotted lattices\xspace}
\newcommand{\parathesaurus}{paraphrase thesaurus\xspace}
\newcommand{\Parathesaurus}{Paraphrase thesaurus\xspace}
\newcommand{\univcd}{\semantics{show-from}} 
\newcommand{\rewrite}{\semantics{rewrite}}

\newcommand{\template}{template\xspace}
\newcommand{\Template}{Template\xspace}
\newcommand{\templates}{templates\xspace}
\newcommand{\Templates}{Templates\xspace}
\renewcommand{\paragraph}[1]{\smallskip \noindent \textbf{#1}~}
\newcommand{\figsnip}{\vspace*{-0.12in}}

\newcommand{\tactic}{lemma\xspace}
\newcommand{\tacticf}{\semantics{lemma100}\xspace}
\newcommand{\tactics}{\semantics{lemma104}\xspace}

\newcommand{\my@node}[3]{%
  \rnode{#1}{\psframebox[framesep=1pt,framearc=.3#3]{#2}}}
\newcommand{\mynode}[2][NONE]{%
  \ifthenelse{\equal{NONE}{#1}}{\my@node{#2}{#2}{}}{\my@node{#1}{#2}{}}}
\newcommand{\myyellownode}[2][NONE]{%
  \ifthenelse{\equal{NONE}{#1}}%
    {\my@node{#2}{#2}{,linestyle=dotted,fillcolor=lightgray,fillstyle=solid}}%
    {\my@node{#1}{#2}{,linestyle=dotted,fillcolor=lightgray,fillstyle=solid}}}
\newcommand{\myboldnode}[2][NONE]{%
  \ifthenelse{\equal{NONE}{#1}}%
    {\my@node{#2}{\textbf{#2}}{,linewidth=2pt}}%
    {\my@node{#1}{\textbf{#2}}{,linewidth=2pt}}}
\newcommand{\myboldyellownode}[2][NONE]{%
  \ifthenelse{\equal{NONE}{#1}}%
    {\my@node{#2}{\textbf{#2}}{,linewidth=2pt,linestyle=dotted,fillcolor=lightgray,fillstyle=solid}}%
    {\my@node{#1}{\textbf{#2}}{,linewidth=2pt,linestyle=dotted,fillcolor=lightgray,fillstyle=solid}}}
\newcommand{\currentlinelabel}{}
\newcommand{\mystartline}[1]{\renewcommand{\currentlinelabel}{#1}}
\newcommand{\mylineto}[2][NONE]{%
  \ifthenelse{\equal{NONE}{#1}}%
    {\ncline{->}{\currentlinelabel}{#2}}%
    {\ncline{->}{#1}{#2}}%
  \mystartline{#2}}
\newcommand{\mydoublelineto}[2][NONE]{%
  \ifthenelse{\equal{NONE}{#1}}%
    {\ncline[offset=1pt]{->}{\currentlinelabel}{#2}%
     \ncline[offset=-1pt]{->}{\currentlinelabel}{#2}}%
    {\ncline[offset=1pt]{->}{#1}{#2}%
     \ncline[offset=-1pt]{->}{#1}{#2}}%
  \mystartline{#2}}
\newcommand{\mytriplelineto}[2][NONE]{%
  \ifthenelse{\equal{NONE}{#1}}%
    {\ncline[offset=0pt]{->}{\currentlinelabel}{#2}%
     \ncline[offset=2pt]{->}{\currentlinelabel}{#2}%
     \ncline[offset=-2pt]{->}{\currentlinelabel}{#2}}%
    {\ncline[offset=0pt]{->}{#1}{#2}%
     \ncline[offset=2pt]{->}{#1}{#2}%
     \ncline[offset=-2pt]{->}{#1}{#2}}%
  \mystartline{#2}}
\newcommand{\myshortcuts}{%
  \newcommand{\N}{\mynode}%
  \newcommand{\yN}{\myyellownode}%
  \newcommand{\bN}{\myboldnode}%
  \newcommand{\byN}{\myboldyellownode}%
  \newcommand{\ybN}{\myboldyellownode}%
  \renewcommand{\S}{\mystartline}%
  \renewcommand{\L}{\mylineto}%
  \newcommand{\dL}{\mydoublelineto}%
  \newcommand{\tL}{\mytriplelineto}
\newcommand{\secgoaltext}{\textit{goal}}
\newcommand{\tactictext}{\textit{lemma}}
}
\makeatother

\begin{document}
\maketitle

\begin{abstract}
An important component of any generation system is the \emph{mapping
dictionary}, a lexicon of elementary semantic expressions and
corresponding natural language realizations.  Typically,
labor-intensive knowledge-based methods are used to construct the
dictionary.  We instead propose to acquire it automatically via a
novel multiple-pass algorithm employing \emph{multiple-sequence
alignment}, a technique commonly used in bioinformatics.  Crucially,
our method leverages latent information contained in \emph{\multip}
corpora --- datasets that supply several verbalizations of the
corresponding semantics rather than just one.
  
We used our techniques to generate natural language versions of
computer-generated mathematical proofs, with good results on both a
per-component and overall-output basis. For example, in evaluations
involving a dozen human judges, our system produced output whose
readability and faithfulness to the semantic input rivaled that of a
traditional generation system.

\end{abstract}

\section{Introduction}
\label{sec:into}

\noindent \emph{One or two homologous sequences whisper \dots a full multiple
  alignment shouts out loud}  \cite{Hubbard+Lesk+Tramontano:96a}.

\smallskip

Today's natural language generation systems typically employ a \emph{lexical
  chooser} that translates complex semantic concepts into words.  The lexical
chooser relies on a \emph{mapping dictionary} that lists possible
realizations of elementary semantic concepts; sample entries might be
\entry{[Parent [sex:female]]}{mother} or
\entrymultiverb{love(\textit{x},\textit{y})}{\textit{x}~loves~\textit{y},
  \textit{x}~is~in~love~with~\textit{y}}.\footnote{ Throughout, fonts denote a
  mapping dictionary's two information types: \semantics{semantics} and
  \realize{realizations}.}

To date, creating these dictionaries has involved human analysis of a
domain-relevant corpus comprised of semantic representations and corresponding
human verbalizations \cite{Reiter+Dale:2000a}.  The corpus analysis and
knowledge engineering work required in such an approach is substantial,
prohibitively so in large domains.  But, since corpus data is already used in
building lexical choosers by hand, an appealing alternative is to have the
system learn a mapping dictionary directly from the data.  Clearly, this would
greatly reduce the human effort involved and ease porting the system to new
domains.  Hence, we address the following problem: given a parallel (but
unaligned) corpus consisting of both complex semantic input and corresponding
natural language verbalizations, learn a semantics-to-words mapping dictionary
automatically.

\setcounter{figure}{1}
\begin{figure*}[tp]
  \begin{center}
    \includegraphics[scale=.4]{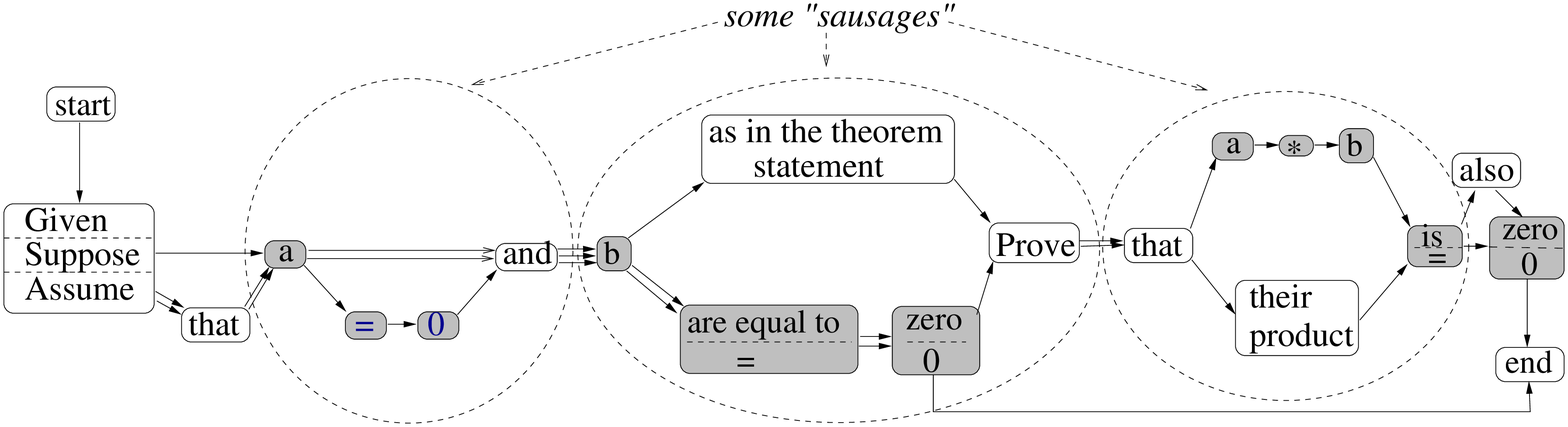}
    \caption{
      Computed \allat for verbalizations from Figure
     \ref{fig:mult-verbs}.  Note how the three indicated ``sausages''
     roughly correspond to the three arguments of
     \univcd(\semantics{a=0,b=0,a$*$b=0}). {\footnotesize (The phrases ``as in the theorem statement'' and ``their product''  correspond to
     chains of nodes, but are drawn as single nodes for
     clarity.  Shading indicates argument-value matches (Section
     \ref{sec:firstphase}). All lattice figures omit punctuation nodes
    for clarity.)}
\figsnip}
     \label{fig:sausage1} \end{center}
\end{figure*}

Now, we could simply apply standard statistical machine translation methods,
treating verbalizations as ``translations'' of the semantics.  These methods
typically rely on \emph{\twop} corpora consisting of text pairs, one in each
``language'' (but cf. \newcite{Simard:99a}; see Section \ref{sec:relwork}).
However, learning the kind of semantics-to-words mapping that we desire from
\twop data alone is difficult even for humans.  First, given the same semantic
input, different authors may (and do) delete or insert information (see Figure
\ref{fig:mult-verbs}); hence, direct comparison between a semantic text and a
single verbalization may not provide enough information regarding their
underlying correspondences.  Second, a single verbalization certainly fails to
convey the variety of potential linguistic realizations of the concept that an
expressive lexical chooser would ideally have access to.

\paragraph{The multiple-sequence idea} Our approach is motivated by
an analogous situation that arises in computational biology.  In brief, an
important bioinformatics problem --- \newcite{Gusfield:97a} refers to it as
``The Holy Grail'' --- is to determine commonalities within a collection of
biological sequences such as proteins or genes.  Because of mutations within
individual sequences, such as changes, insertions, or deletions, pair-wise
comparison of sequences can fail to reveal which features are conserved across
the entire group.  Hence, biologists {compare multiple sequences
  simultaneously} to reveal hidden structure characteristic to the group as a
whole.

\emph{Our work applies multiple-sequence alignment techniques to the
  mapping-dictionary acquisition problem}.  The main idea is that using a
\emph{\multip corpus} --- one that supplies \textit{several} alternative
verbalizations for each semantic expression --- can enhance both the accuracy
and the expressiveness of the resulting dictionary.  In particular, matching a
semantic expression against a composite of the common structural features of a
set of verbalizations ameliorates the effect of ``mutations'' within individual
verbalizations.  Furthermore, the existence of multiple verbalizations
helps the system learn several ways to express concepts.

To illustrate, consider a sample semantic expression from the mathematical
theorem-proving domain.  The expression \univcd(\semantics{a=0,b=0,a$*$b=0})
means ``assuming the two premises $a=0$ and $b=0$, show that the goal $a*b=0$
holds''. Figure \ref{fig:mult-verbs} shows three human verbalizations of this
expression.  \setcounter{figure}{0}
\begin{figure}[t]
\begin{center}
{\footnotesize
  \begin{frameit}{0.9\columnwidth}
    \hline
    (1) Given a and b as in the theorem statement,\\ prove that {a$*$b=0}. \\
    \hline
    (2) Suppose that a and b are equal to zero. \\
    Prove that their product is also zero. \\
    \hline
    (3) Assume that {a=0} and {b=0}.
  \end{frameit}
  }
\end{center}
\caption{\label{fig:mult-verbs}
Three different
human verbalizations of
\univcd(\semantics{a=0,b=0,a$*$b=0}).\figsnip}
\end{figure}
Even for so formal a domain as mathematics, the verbalizations vary
considerably, and none directly matches the entire semantic input.  For
instance, it is not obvious without domain knowledge that ``Given $a$ and $b$
as in the theorem statement'' matches ``\semantics{a=0}'' and
``\semantics{b=0}'', nor that ``their product'' and ``\semantics{a$*$b}'' are
equivalent. Moreover, sentence (3) omits the goal argument entirely.  However,
as Figure~\ref{fig:sausage1} shows, the combination of these verbalizations, as
computed by our multiple-sequence alignment method, exhibits high structural
similarity to the semantic input: the indicated ``sausage'' structures
correspond closely to the three arguments of \univcd.

\section{Multiple-sequence alignment}
\label{sec:msa}

This section describes general multiple-sequence alignment; we discuss its
application to learning mapping dictionaries in the next section.

A multiple-sequence alignment algorithm takes as input $n$ strings and outputs
an $n$-row correspondence table, or \emph{multiple-sequence alignment} (MSA).
(We explain how the correspondences are actually computed below.)  The \msa's
rows correspond to sequences, and each column indicates which elements of which
strings are considered to correspond at that point; non-correspondences, or
``gaps'', are represented by underscores ($\underscore$).  See
Figure~\ref{fig:msa}(i).

\setcounter{figure}{2}
\begin{figure}[hbtp]
  \begin{center}
    \includegraphics[scale=.5]{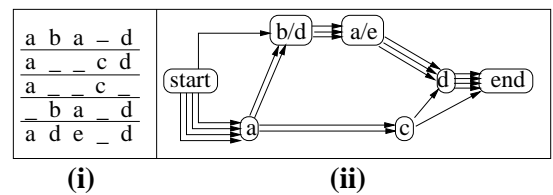}
    \caption{(i) An \msa  of five sequences
 (the first  is ``abad''); (ii) The corresponding \allat. \figsnip}
    \label{fig:msa}
  \end{center}
\end{figure}

From an \msa, we can compute a \emph{\allat}.  Each \allat node, except for
``start'' and ``end'', corresponds to an \msa column.  The edges are induced by
traversing each of the \msa's rows from left to right. See
Figure~\ref{fig:msa}(ii).

\paragraph{Alignment computation} 
The sum-of-pairs dynamic programming algorithm and pairwise iterative alignment
algorithm sketched here are described in full in \newcite{Gusfield:97a} and
Durbin et al.~\shortcite{Durbin+al:98a}.

Let $\alphabet$ be the set of elements making up the sequences to be aligned,
and let $\simfn(x,y)$, $x \mbox{ and } y \in \Sigma \cup \set{\underscore}$, be
a domain-specific \emph{similarity function} that assigns a score to every
possible pair of alignment elements, including gaps.  Intuitively, we prefer
\msas in which many high-similarity elements are aligned.

In principle, we can use dynamic programming over alignments of sequence
prefixes to compute the highest-scoring \msa, where the \emph{sum-of-pairs}
score for an \msa is computed by summing $\simfn(x,y)$ over each pair of
entries in each column.  Unfortunately, these computations are exponential in
$n$, the number of sequences. (In fact, finding the optimal \msa when $n$ is a
variable is NP-complete \cite{Wang+Jiang:94a}.)  Therefore, we use
\emph{iterative pairwise} alignment, a commonly-used polynomial-time
approximation procedure, instead.  This algorithm greedily merges pairs of
\msas of (increasingly larger) \emph{subsets} of the $n$ sequences; which pair
to merge is determined by the average score of all pairwise alignments of
sequences from the two \msas.

\paragraph{Aligning \allats}
We can apply the above sequence alignment algorithm to \allats as well as
sequences, as is indeed required by pairwise iterative alignment.  We simply
treat each \allat as a sequence whose $i$th symbol corresponds to the {set} of
nodes at distance $i$ from the start node.  We modify the similarity function
accordingly: any two new symbols are equivalent to subsets $S_1$ and $S_2$ of
$\alphabet$, so we define the similarity of these two symbols as $\max_{(x,y)
  \in S_1 \times S_2} \simfn(x,y)$.

\section{Dictionary Induction}
\label{sec:learning}

Our goal is to produce a semantics-to-words mapping dictionary by comparing
semantic sequences to \msas of multiple verbalizations.  We assume only that
the semantic representation uses predicate-argument structure, so the
elementary semantic units are either \emph{terms} (e.g., $\semantics{0}$), or
\emph{predicates} taking arguments (e.g., $\univcd(prem1,prem2,goal)$, whose
arguments are two premises and a goal).  Note that both types of units can be
verbalized by multi-word sequences.

Now, semantic units can occur several times in the corpus.  In the case of
predicates, we would like to combine information about a given predicate from
\emph{all} its appearances, because doing so would yield more data for us to
learn how to express it.  On the other hand, correlating verbalizations across
instances instantiated with different argument values (e.g.,~
$\semantics{\univcd(a=0,b=0,a*b=0)}$ vs. $\semantics{\univcd(c>0,d>0,c/d>0)}$)
makes alignment harder, since there are fewer obvious matches (e.g.,
``a$*$b=0'' does not greatly resemble ``{c/d$>$0}''); this seems to discourage
aligning cross-instance verbalizations.

We resolve this apparent paradox by a novel three-phase approach:
\begin{itemize}
\item In the \emph{\firstphase} phase (Section \ref{sec:firstphase}), we handle
  each separate instance of a semantic predicate individually.  First, we
  compute a separate \msa for each instance's verbalizations.  Then, we
  abstract away from the particular argument values of each instance by
  replacing \allat portions corresponding to argument values with
  \emph{argument slots}, thereby creating a {\em \slotlat}.
\item In the \emph{\secondphase} phase (Section \ref{sec:secondphase}), for
  each predicate we align together {all} the {\slotlats} from \emph{all} of its
  instances.
\item In the \emph{\template induction} phase (Section \ref{sec:templates}), we
  convert the aligned \slotlats into \emph{\templates} --- sequences of words
  and argument positions --- by tracing \slotlat paths.
\end{itemize}
Finally, we enter the \templates into the mapping dictionary.

\subsection{\Firstphase}
\label{sec:firstphase}

As mentioned above, the first job of the \firstphase phase is to separately
compute for each instance of a semantic unit an \msa of all its verbalizations.
To do so, we need to supply a scoring function capturing the similarity in
meaning between words.  Since such similarity can be domain-dependent, we use
the data to induce --- again via sequence alignment --- a \emph{\parathesaurus}
$T$ that lists linguistic items with similar meanings.  (This process is
described later in section \ref{sec:paraphrase}.)  We then set
\newcommand{\alignalphabet}{\alphabet}
$$\simfn(x,y) = \cases{ 1 & $x = y$, $x \in \alignalphabet$; \cr 0.5 & $x
  \approx y$; \cr -0.01 & exactly one of $x,y$ is $\underscore$~; \cr -0.5 &
  otherwise (mismatch)}
$$
where $\alphabet$ is the vocabulary and $x \approx y$ denotes that $T$ lists
$x$ and $y$ as paraphrases.\footnote{These values were hand-tuned on a held-out
  development corpus, described later.  Because we use progressive alignment,
  the case $x = y = \underscore$ does not occur.}  Figure~\ref{fig:sausage1}
shows the \allat computed for the verbalizations of the instance
\semantics{\univcd(a=0,b=0,a$*$b=0)} listed in Figure \ref{fig:mult-verbs}.
The structure of the \allat reveals why we informally refer to \allats as
``sausage graphs''.

\begin{figure}[bhtp]
  \begin{center}
    \includegraphics[scale=.4]{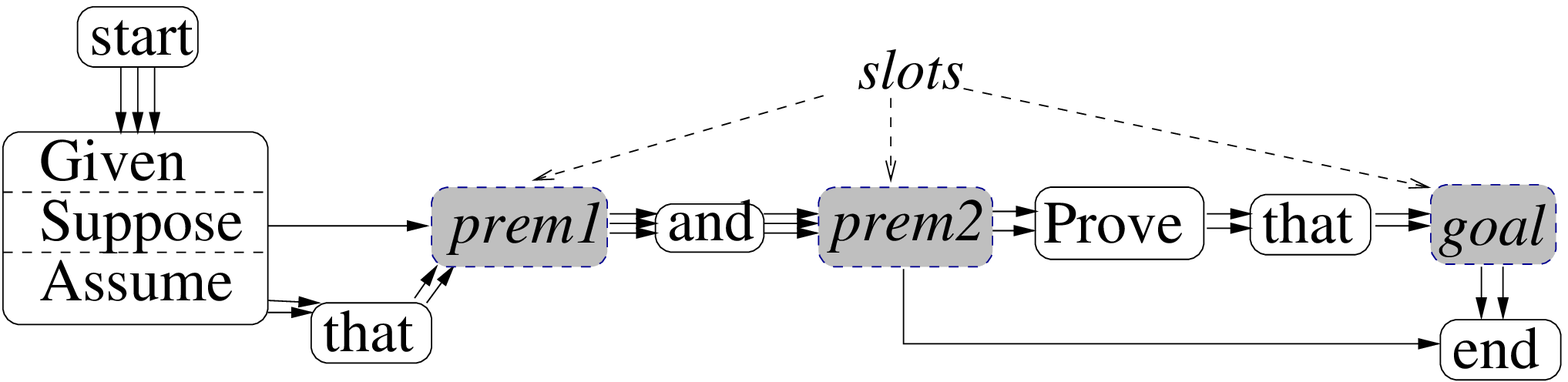}
    \caption{
\Slotlat, computed from the \allat in Figure \ref{fig:sausage1}, for $\univcd$$(prem1,prem2,goal)$. \vspace*{-.2in}}
    \label{fig:sausage1-holed}
  \end{center}
\end{figure}

Next, we transform the \allats into \slotlats.  We use a simple matching
process that finds, for each argument value in the semantic expression, a
sequence of \allat nodes such that each node contains a word identical to or a
paraphrase of (according to the \parathesaurus) a symbol in the argument value
(these nodes are shaded in Figure \ref{fig:sausage1}).  The sequences so
identified are replaced with a ``slot'' marked with the argument variable (see
Figure \ref{fig:sausage1-holed}).\footnote{This may further change the topology
  by forcing other nodes to be removed as well.  For example, the \slotlat in
  Figure~\ref{fig:sausage1-holed} doesn't contain the node sequence ``their
  product''.}  Notice that by replacing the argument values with variable
labels, we make the commonalities between \slotlats for different instances
more clear, which is useful for the \secondphase step.

\subsubsection{\Parathesaurus creation}
\label{sec:paraphrase}

Recall that the \parathesaurus plays a role both in aligning verbalizations and
in matching \allat nodes to semantic argument values.  The main idea behind our
\parathesaurus induction method, motivated by \newcite{Barzilay+McKeown:01a},
is that paths through \allat ``sausages'' often correspond to alternate
verbalizations of the same concept, since the sausage endpoints are contexts
common to all the sausage-interior paths.  Hence, to extract paraphrases, we
first compute all {pairwise} alignments of parallel verbalizations, discarding
those of score less than four in order to eliminate spurious matches.\footnote{
  Pairwise alignments yield fewer candidate alignments from which to select
  paraphrases, allowing simple scoring functions to produce decent results.}
Parallel sausage-interior paths that appear in several alignments are recorded
as paraphrases.  Then, we iterate, realigning each pair of sentences, but with
previously-recognized paraphrases treated as identical, until no new
paraphrases are discovered.  While the majority of the derived paraphrases are
single words,  the algorithm also produces several multi-word paraphrases,
such as ``are equal to'' for ``=''.
To simplify subsequent comparisons, these phrases
(e.g., ``are equal to'') are treated as single tokens.
Here are four paraphrase pairs we extracted from the 
mathematical-proof domain: 
\begin{center}
  {\footnotesize
    \begin{tabular}{|l|l|}
      \hline
      (conclusion, result) &  (0, zero)     \\
      (applying, by)  &    (expanding, unfolding)       \\
      \hline
    \end{tabular}
    }
\end{center}
(See Section \ref{sec:eval-comp} for a formal evaluation of the paraphrases.)
We treat thesaurus entries as degenerate \slotlats containing no slots;
hence, terms and predicates are represented in the same way.

\begin{figure*}[t]
  \begin{center}
  \fbox{\parbox{0.8\textwidth}{\myshortcuts\scriptsize\noindent
  \centerline{\N{Then} ~ \N{we} ~ \N{can} ~ \N{use} ~ \N{lemma} ~
  \yN[tac]{\tactictext} ~ \N[formula]{$\frac{a}{n}=\frac{-a}{-n}$} ~
  \N{and} ~ \N{get} ~ \yN[goal2]{\secgoaltext}} \\[1ex]
  \centerline{\N{start} \hfill \N{end}} \\[1ex] \centerline{~~~~~~
  \N{Now} ~~~~~~~~~~~~ \N[the1]{the} ~ \N{fact} ~ \N{about} ~
  \N{division} ~~~~ \N{to} ~~~~~~~~ \N[the2]{the} ~ \N{goal}}
  \L[start]{Then}\L{we}\L{can}\L{use}\L{lemma}\L{tac}\L{formula}\L{and}
  \L{get}\L{goal2}\dL{end}
  \L[start]{Now}\L{use}\L{the1}\L{fact}\L{about}\L{division}\L{formula}
  \L{to}\L{get}\L{the2}\L{goal}\L{goal2} \\[1ex] \centerline{\N{we} ~
  \N{can} ~ \N[useapp]{$\frac{\mbox{use}}{\mbox{apply}}$} ~~~
  \N{lemma} ~~~ \yN[tac]{\tactictext} ~~~ \N{to} ~ \N{get} ~
  \yN[goal2]{\secgoaltext}} \\[1ex] \centerline{\N{start} \hfill
  \N{end}} \\[1ex] \centerline{~~~~~~ \N{then} \hspace{20em} \N{the} ~
  \N[left]{left-hand} ~ \N{side}}
  \L[start]{we}\L{can}\L{useapp}\dL{lemma}\dL{tac}\dL{to}\L{get}\L{goal2}
  \L{end} \L[start]{then}\L{we}
  \nccurve[angle=-90,linearc=.2]{->}{we}{useapp}
  \L[to]{the}\L{left}\L{side}\L{end}}} \end{center} \caption{\Slotlats
  for the predicate \semantics{\rewrite}\textit{(lemma,goal)} derived
  from two instances: \\
\hspace*{.2in}(instance I)
  \semantics{\rewrite(\tacticf,a-n*((-a)/(-n))=-(-a-(-n)*((-a)/(-n))))}, and \\ 
\hspace*{.2in}(instance II)
\semantics{\rewrite(\tactics,A-(-(A/(-N)))*N  = A-(A/(-N))*(-N))}; \\
each instance had two verbalizations.
In instance (I), both verbalizations
contain  the context-dependent information
``$\frac{a}{n}=\frac{-a}{-n}$'' (the statement of \tacticf); also,
argument-matching failed on the context-dependent phrase ``the fact about division''.  
}
 \label{fig:rewrite-2}
\end{figure*}

\begin{figure*}[t]
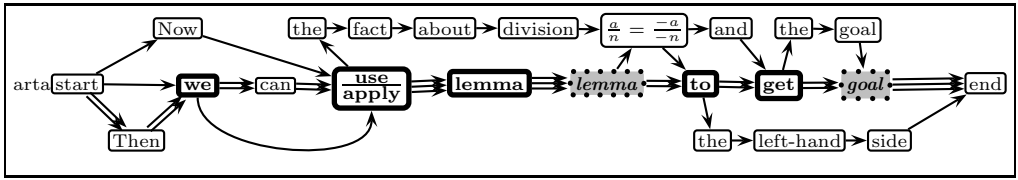

  \begin{center}
  \fbox{\parbox{0.8\textwidth}{\myshortcuts\scriptsize\noindent
      \centerline{~ \N{Now} ~ \hspace{2em}
                  ~ \N[the1]{the} ~ \N{fact} ~ \N{about} ~ \N{division}
                  ~ \N[formula]{$\frac{a}{n}=\frac{-a}{-n}$}
                  ~ \N{and} ~ \N[the2]{the} ~ \N{goal}}
      \\[2ex]
      \centerline{\N{{start}}
                  \hfill \bN{we}
                  ~~~ \N{can}
                  ~~~ \bN[useapp]{$\frac{\mbox{use}}{\mbox{apply}}$}
                  ~~~ \bN{lemma}
                  ~~~ \byN[tac]{\tactictext}
                  ~~~ \bN{to}
                  ~~~ \bN{get}
                  ~~~ \byN[goal2]{\secgoaltext}
                  \hfill \N{end}}
      \\[2ex]
      \centerline{\N{Then} ~ \hspace{20em}
                  ~ \hspace{2em}
                  ~ \N{the} ~ \N[left]{left-hand} ~ \N{side}}
      \dL[start]{Then}\dL{we}
      \nccurve[angle=-90,linearc=.2]{->}{we}{useapp}
      \L[tac]{formula}\L{and}\L{get}
      \L[formula]{to}\L{the}\L{left}\L{side}\L{end}
      \L[start]{we}\dL{can}\dL{useapp}\tL{lemma}\tL{tac}\dL{to}\dL{get}
      \dL{goal2}\tL{end}
      \L[start]{Now}\L{useapp}\L{the1}\L{fact}\L{about}\L{division}\L{formula}
      \L[get]{the2}\L{goal}\L{goal2}}}

    \caption{Unified \slotlat computed by \secondphase of
      Figure \ref{fig:rewrite-2}'s \slotlats.  The consensus sequence is
      shown in bold (recall that node weight roughly
      corresponds to in-degree).\figsnip }  
    \label{fig:rewrite-unif}
    \end{center}
\end{figure*}

\subsection{\Secondphase}
\label{sec:secondphase}

Figure \ref{fig:sausage1-holed} 
is an example where
the
verbalizations for a single instance yield good information as to how
to realize a predicate.
(For example, ``\realize{Assume} \textit{[prem1]} \realize{and}
\textit{[prem2]}, \realize{prove} \textit{[goal]}'', where the brackets enclose arguments marked with their type.) Sometimes, though, the
situation is more complicated. Figure~\ref{fig:rewrite-2} shows two \slotlats
for different instances of $\rewrite(\tactic,goal)$ (meaning, rewrite
\textit{goal} by applying $\tactic$); the first \slotlat is problematic because
it contains context-dependent information (see caption).  Hence, we engage in
\emph{\secondphase} to merge information about the predicate.  That is, we
align the \slotlats for \emph{all} instances of the predicate (see
Figure~\ref{fig:rewrite-unif}); the resultant \emph{unified \slotlat} reveals
linguistic expressions common to verbalizations of different instances.  Notice
that the argument-matching process in the \firstphase phase helps make these
commonalities more evident by abstracting over different values of the same
argument (e.g., \tacticf and \tactics are both relabeled ``$\tactic$'').

\subsection{\Template\ induction}
\label{sec:templates}

Finally, it remains to create the mapping dictionary from unified \slotlats.
While several strategies are possible, we chose a simple \emph{consensus
  sequence} method.  Define the \emph{node weight} of a given \slotlat node as
the number of verbalization paths passing through it (downweighted if it
contains punctuation or the words ``the'', ``a'', ``to'', ``and'', or ``of'').
The \emph{path weight} of a \slotlat path is a length-normalized sum of the
weights of its nodes.\footnote{Shorter paths are preferred, but we discard
  sequences shorter than six words as potentially spurious.}  We produce as a
\template the words from the \emph{consensus sequence}, defined as the
maximum-weight path, which is easily computed via dynamic programming.  For
example, the \template we derive from Figure~\ref{fig:rewrite-unif}'s \slotlat
is \realize{We use lemma} [{\em \tactic}] \realize{to get} [\textit{goal}].

While this method is quite efficient, it does not fully exploit the
expressive power of the lattice, which may encapsulate several valid realizations.
We leave to future work experimenting with alternative
\template-induction techniques; see Section \ref{sec:relwork}.

\begin{figure*}
  \begin{center}
\begin{tabular}{|c|c|c|}\hline
(i) \omt{Nuprl} & (ii) \omt{\msa-based output} & (iii) \omt{Traditional output}\\\hline

\begin{minipage}{2.7in}
{\footnotesize
        \semantics{UnivCD($\forall$ i:$\mathbb{N}$.$|$\semantics{i}$|=|$\semantics{-i}$|$, i:$\mathbb{N}$},
          $|$\semantics{i}$|=|$\semantics{-i}$|$)\\
        \semantics{BackThruLemma}($|$\semantics{i}$|=|$\semantics{-i}$|$, \semantics{i}= $\pm$ \semantics{i},\semantics{absval{\underscore}eq})\\
        \semantics{Unfold(i= $\pm$
          i,}~~~\semantics{(),}~~~\semantics{pm{\underscore}equal)}
\vspace*{.3in}
        }\end{minipage}
&
\begin{minipage}{1.55in}
\smallskip
{\footnotesize
        Assume that $i$ is an integer, we need to show
        $|i| = |-i|$.  From \texttt{absval\_eq} lemma, $|i| = |-i|$ reduces
        to $i = \pm i$.  By the definition of \texttt{pm\_equal}, $i = \pm i$
        is proved.\\
        }
\end{minipage}
&
\begin{minipage}{1.55in}
  {\footnotesize Assume $i$ is an integer. By the \texttt{absval\_eq} lemma,
    the goal becomes $|i| = |-i|$.  Now, the original expression can be
    rewritten as $i = \pm i$.  \vspace*{.1in} }
\end{minipage} \\ \hline
\end{tabular}
    \caption{(i) Nuprl proof 
(test lemma ``h'' in Figure \ref{fig:read}).
      (ii) Verbalization produced by our system.  (iii) Verbalization
    produced by traditional generation system; note that the initial
    goal is never specified, which means that in the
    phrase ``the
    goal becomes'', we don't know what the goal is.\figsnip}
    \label{fig:proof}
  \end{center}
\end{figure*}

\section{Evaluation}

We implemented our system on formal mathematical proofs created by the Nuprl
system, which 
has been used to create thousands of proofs in many
mathematical fields \cite{Constable:86a}.  Generating natural-language versions
of proofs was first addressed several decades ago~\cite{Chester:76a}.  But now,
large formal-mathematics libraries are available on-line.\footnote{See
  \texttt{http://www.cs.cornell.edu/\-Info/\-Projects/\-NuPrl/} or
  \texttt{http://www.mizar.org}, for example.}  Unfortunately, they are usually
encoded in highly technical languages (see Figure~\ref{fig:proof}(i)).
Natural-language versions of these proofs would make them more widely
accessible, both to users lacking familiarity with a specific prover's
language, and to search engines which at present cannot search the symbolic
language of formal proofs.

Besides these practical benefits, the formal mathematics domain has the further
advantage of being particularly suitable for applying statistical generation
techniques.  Training data is available because theorem-prover developers
frequently provide verbalizations of system outputs for explanatory purposes.
In our case, a \multip corpus of Nuprl proof verbalizations already exists
\cite{Holland-Minkley+Barzilay+Constable:99a} and forms the core of our
training corpus.  Also, from a research point of view, the examples from Figure
\ref{fig:mult-verbs} show that there is a surprising variety in the data,
making the problem quite challenging.

All evaluations reported here involved judgments from graduate students and
researchers in computer science.
We authors were not among the judges.

\subsection{Corpus}
\label{sec:corpus}

Our training corpus consists of 30 Nuprl proofs and 83 verbalizations.  On
average, each proof consists of 5.08 \emph{proof steps}, which are the basic
semantic unit in Nuprl; Figure~\ref{fig:proof}(i) shows an example of three
Nuprl steps.  An additional five proofs, disjoint from the test data, were used
as a development set for setting the values of all parameters.\footnote{ See
  \texttt{http://www.cs.cornell.edu/Info/Projects/\\NuPrl/html/nlp} for all our
  data.}  \newcommand{\showterms}[1]{#1}

\smallskip
\noindent \textbf{Pre-processing}~
First, we need to divide the verbalization texts into portions corresponding to
individual proof steps, since \firstphase handles verbalizations for only one
semantic unit at a time.  Fortunately,
\newcite{Holland-Minkley+Barzilay+Constable:99a} showed that for Nuprl, one
proof step roughly corresponds to one sentence in a
natural language verbalization.  So, we align Nuprl steps with verbalization
sentences using dynamic programming based on the number of symbols common to
both the step and the verbalization.  This produced 382 pairs of Nuprl steps
and corresponding verbalizations.  We also did some manual cleaning on the
training data to reduce noise for subsequent stages.\footnote{We employed
  pattern-matching tools to fix incorrect sentence boundaries, converted
  non-ascii symbols to a human-readable format, and discarded a few
  verbalizations which were unrelated to the underlying proof.}

\subsection{Per-component evaluation}
\label{sec:eval-comp}

We first evaluated three individual components of our system: \parathesaurus
induction, argument-value selection in \slotlat induction, and \template
induction.  We also validated the utility of \multip, as opposed to \twop,
data.

\paragraph{\Parathesaurus}
We presented
two 
judges with all 71 paraphrase pairs produced by our system.  They identified
87\% and 82\%, respectively, as being plausible substitutes within a
mathematical context.

\paragraph{Argument-value selection}
We next measured how well our system matches semantic argument values with
\allat node sequences.  We randomly selected 20 Nuprl steps and their
corresponding verbalizations.  From this sample, a Nuprl expert identified the
argument values that appeared in at least one corresponding verbalization; of
the 46 such values, our system correctly matched \allat node sequences to
91\%.  To study the relative effectiveness of using \multip rather than
\twop data, we also implemented a baseline system that used only \emph{one}
(randomly-selected) verbalization among the multiple possibilities.
This single-verbalization baseline matched only 44\% of the values
correctly,
indicating the value of a \multip-corpus approach.

\paragraph{\Templates}
Thirdly, we randomly selected 20 induced \templates; of these, a Nuprl expert
determined that 85\% were plausible verbalizations of the corresponding Nuprl.
This was a very large improvement over the single-verbalization baseline's
30\%, again validating the \multip-corpus approach.

\newcommand{\A}{&$\blacksquare$}
\newcommand{\B}{&$\square$} \newcommand{\X}{&$\blacklozenge$}
\newcommand{\Q}{&?}  \newcommand{\Asymb}{$\blacksquare$}
\newcommand{\Bsymb}{$\square$} \newcommand{\Xsymb}{$\blacklozenge$}
\newcommand{\Y}{$\checkmark$}
\begin{figure*}
\begin{center}
{\small

\begin{tabular}{|l||c@{~}c@{~}c@{~}c@{~}c@{~}c@{~}c@{~}c@{~}c@{~}c@{~}c@{~}c@{~}c@{~}c@{~}c@{~}c@{~}c@{~}c@{~}c@{~}c|r|}
      \hline
        & \multicolumn{20}{|c|}{Lemma} &  \% good  \\
Judge    &a &b &c &d &e  &f &g &h &i &j &k &l &m &n &o &p &q &r &s &t&
      \\\hline
      A \A \A \A \A \X \A \A \A \A \A \A \X \A \A \A \A \A \X \A \X &100\\
      B \B \B \A \B \A \A \X \A \A \X \X \X \B \A \A \A \A \A \B \A &75 \\
      C \X \X \B \B \X \A \B \A \B \B \A \X \X \A \A \X \B \A \X \X &70 \\
      D \B \A \A \X \X \X \A \B \A \B \B \A \X \B \A \A \B \A \X \A &70 \\
      E \B \X \A \X \X \X \A \X \A \B \B \X \X \B \X \B \A \X \B \X &70 \\
      F \A \A \A \A \B \X \B \A \A \A \A \A \A \A \B \A \A \A \A \A &85 \\
      G \B \A \A \A \X \B \A \A \X \A \A \X \A \A \B \X \X \A \X \X &85 \\
      H \A \A \A \A \A \A \A \A \A \A \A \A \A \A \A \A \A \A \A \A &100\\
      I \B \B \A \X \X \B \B \A \A \B \A \B \B \A \X \X \A \A \A \B &60 \\
      J \X \X \A \X \A \X \X \A \X \X \X \X \A \A \A \A \B \B \B \A &85 \\
      K \A \A \A \B \A \A \A \B \B \X \B \B \A \A \A \A \B \B \X \A
        &65 \\ \hline
\% good &55 &82 &91 &73 &91 &82 &73 &82 &82 &64 &73 &82 &82 &82 &82
        &91 &64 &82 &73 &91 &  \\ \hline
$> 50\%$ \Asymb? &   &\Y &\Y &  &  &  &\Y &\Y &\Y &  &\Y &  &\Y &\Y
      &\Y &\Y &\Y &\Y &  &\Y &  \\
      \hline
    \end{tabular}
}
\end{center}
\caption{\label{fig:read} Readability results.  \Asymb: preference
for our system.  \Bsymb: preference for hand-crafted
system.  \Xsymb: no preference.  $\checkmark$: $> 50\%$ of the judges
preferred the statistical system's output.\figsnip}
\end{figure*}

\subsection{Evaluation of the generated texts}
Finally, we evaluated the quality of the text our system generates by comparing
its output against the system of
\newcite{Holland-Minkley+Barzilay+Constable:99a}, which produces accurate and
readable Nuprl proof verbalizations.  Their system has a hand-crafted lexical
chooser derived via manual analysis of the same corpus that our system was
trained on.  To run the experiments, we replaced Holland-Minkley et. al's
lexical chooser with the mapping dictionary we induced.  (An alternative
evaluation would have been to compare our output with human-authored texts.
But this wouldn't have allowed us to evaluate the performance of the lexical
chooser alone, as human proof generation may differ in aspects other than
lexical choice.)  The test set serving as input to the two systems consisted of
20 held-out proofs, unseen throughout the entirety of our algorithm development
work.  We evaluated the texts on two dimensions: readability and fidelity to
the mathematical semantics.

\paragraph{Readability} We asked 11 judges to compare the readability of the
texts produced from the same Nuprl proof input: Figure \ref{fig:proof}(ii) and
(iii) show an example text pair.\footnote{To prevent judges from identifying
  the system producing the text, the order of presentation of the two systems'
  output was randomly chosen anew for each proof.}  (The judges were not given
the original Nuprl proofs.)  Figure \ref{fig:read} shows the results.
\emph{Good} entries are those that are \emph{not} preferences for the
traditional system, since our goal, after all, is to show that \msa-based
techniques can produce output as good or better than a hand-crafted system.  We
see that for \emph{every} lemma and for \emph{every} judge, our system
performed quite well.  Furthermore, for more than half of the lemmas, more than
half the judges found our system's output to be distinctly better than the
traditional system's.

\paragraph{Fidelity} 
We asked a Nuprl-familiar expert in formal logic to
determine, given the Nuprl proofs and output texts,  whether the texts preserved the main ideas of the formal proofs
without introducing ambiguities.  All 20 of our system's proofs were judged
correct, while only 17 of the traditional system's proofs were judged to be
correct.

\section{Related Work}
\label{sec:relwork}

Nuprl creates proofs at a higher level of abstraction than other
provers
do, so we were able to learn verbalizations directly
from the Nuprl proofs themselves.  In other natural-language proof generation
systems \cite{Huang+Fiedler:97a,Siekmann:99a} and other generation
applications, the semantic expressions to be realized are the product of the
system's content planning component, not the proof or data.  But our techniques
can still be incorporated into such systems, because we can
map verbalizations to the content planner's output.  Hence, we believe our
approach generalizes to other settings.

Previous research on statistical generation has addressed different problems.
Some systems learn from verbalizations annotated with semantic concepts
\cite{Ratnaparkhi:00a,Oh+Rudnicky:00a}; in contrast, we use un-annotated
corpora.  Other work focuses on \emph{surface realization} --- choosing among
different lexical and syntactic options supplied by the lexical chooser and
sentence planner --- rather than on creating the mapping dictionary; although
such work also uses lattices as input to the stochastic realizer, the lattices
themselves are constructed by traditional knowledge-based means
\cite{Langkilde+Knight:98a,Bangalore+Rambow:00a}.  An exciting direction for
future research is to apply these statistical surface realization methods to
the lattices our method produces.

Word lattices are commonly used in speech recognition to represent different
transcription hypotheses.  \newcite{Mangu+Brill+Stolcke:00a} compress these
lattices into \emph{confusion networks} with structure reminiscent of our
``sausage graphs'', utilizing alignment criteria based on word identity and
external information such as phonetic similarity.

Using alignment for grammar and lexicon induction has been an active area of
research, both in monolingual settings \cite{vanZaanen:00a} and in machine
translation (MT) \cite{Brownetal:93a,Melamed:00a,Och+Ney:00a} ---
interestingly, statistical MT techniques have been used to derive
lexico-semantic mappings in the ``reverse'' direction of language
\emph{understanding} rather than generation
\cite{Papineni+Roukos+Ward:97a,Macherey+Och+Ney:01a}.  
In a preliminary study, applying IBM-style alignment models
in a black-box manner (i.e., without modification) to our setting did
not yield promising results \cite{Chong:02a}.  On the other hand, MT
systems can often model \emph{crossing} alignment situations; these
are rare in our data, but we hope to account for them in future work.

While recent proposals for \emph{evaluation} of MT systems have
involved \multip corpora \cite{Thompson+Brew:96a,Papineni+al:02a},
statistical MT \emph{algorithms} typically only use \twop data.
Simard's \shortcite{Simard:99a} \emph{trilingual} (rather than
\multip) corpus method, which also computes \msas, is a notable
exception, but he reports mixed experimental results.  In contrast, we
have shown that through application of a novel composition of
alignment steps, we can leverage \multip corpora to create
high-quality mapping dictionaries supporting effective text
generation.

{\small 
\section*{Acknowledgments}
We thank Stuart Allen, Eli Barzilay, Stephen Chong, Michael Collins,
Bob Constable,  Jon
Kleinberg, John Lafferty, Kathy McKeown,  Dan Melamed, Golan Yona, the Columbia NLP group, and
the anonymous reviewers for many helpful comments.  
Thanks also to the Cornell
Nuprl and Columbia NLP groups, Hubie Chen, and Juanita Heyerman for
participating in our evaluation, and the Nuprl group for 
generating verbalizations.  We are grateful to Amanda Holland-Minkley
for help running the comparison experiments.  Portions of this work
were done while the first author was visiting Cornell University. This
paper is based upon work supported in part by the National Science
Foundation under ITR/IM grant IIS-0081334 and a Louis Morin
scholarship. Any opinions, findings, and conclusions or
recommendations expressed above are those of the authors and do not
necessarily reflect the views of the National Science Foundation.

\bibliographystyle{acl}

}
\end{document}